\documentclass{article}

\usepackage[utf8]{inputenc} % allow utf-8 input
\usepackage[T1]{fontenc}
\usepackage{hyperref}
\usepackage{url}
\usepackage{booktabs}
\usepackage{amsfonts}
\usepackage{nicefrac}
\usepackage{microtype}
\usepackage[table,dvipsnames]{xcolor}

\PassOptionsToPackage{numbers, compress}{natbib}

\usepackage[preprint]{neurips_2025}

\usepackage{graphicx}
\usepackage{subcaption}
\usepackage{float}
\usepackage[justification=justified]{caption}
\usepackage{lscape}
\usepackage{epsfig}
\usepackage{wrapfig}
\usepackage{svg}

\usepackage[lined,ruled,linesnumbered]{algorithm2e}

\usepackage{booktabs}
\usepackage{multirow}
\usepackage{makecell}
\usepackage{array}

\usepackage{paralist}
\usepackage{enumitem}

\usepackage{bm}
\usepackage{mathtools}
\usepackage{amssymb,amsmath}

\usepackage{units}
\usepackage{color}
\usepackage{pifont}
\usepackage{newtxtext}

\usepackage{xspace}

\usepackage{comment}

\usepackage[capitalize]{cleveref}
\crefname{section}{Sec.}{Secs.}
\Crefname{section}{Section}{Sections}
\Crefname{table}{Table}{Tables}
\crefname{table}{Tab.}{Tabs.}

\def\etal{\textit{et~al.}}
\def\eg{e.g.,~}

\def\vs{vs.~}

\newlength\paramarginsize
\newlength\figmarginsize
\newlength\secmarginsize
\newlength\figcapmarginsize
\newlength\tabcapmarginsize

\newcommand{\figcapmargin}{\vspace{\figcapmarginsize}}
\newcommand{\tabcapmargin}{\vspace{\tabcapmarginsize}}

\newcommand{\mpage}[2]
{
\begin{minipage}{#1\linewidth}\centering
#2
\end{minipage}
}

\newcommand{\figcaption}[2]
{
\caption{
\textbf{#1.}
#2
}
}

\newcommand{\secref}[1]{Section~\ref{sec:#1}}
\newcommand{\figref}[1]{Figure~\ref{fig:#1}}
\newcommand{\tabref}[1]{Table~\ref{tab:#1}}
\newcommand{\eqnref}[1]{\eqref{eq:#1}}

\long\def\ignorethis#1{}

\newcommand{\cmark}{\ding{51}}%
\newcommand{\xmark}{\ding{55}}%

\definecolor{gradA}{HTML}{E9E3F9}
\definecolor{gradB}{HTML}{DDE9FF}
\definecolor{gradC}{HTML}{D8F5F4}
\definecolor{gradD}{HTML}{DDF8EE}

\definecolor{gradmid}{RGB}{224,238,243}

\definecolor{mygray}{gray}{0.9}
\definecolor{myred}{HTML}{ffaaaa}

\definecolor{mygreen}{HTML}{bbffbb}

\def\xi{\mathbf{x}_i}

\def\ours{Flashback}
\graphicspath{{figure}, {images}, {example}}

\title{Flashback: Memory-Driven Zero-shot, Real-time Video Anomaly Detection}

\author{%
  Hyogun Lee,
  Haksub Kim,
  Ig-Jae Kim,
  Yonghun Choi\thanks{Corresponding author} , \\
  Korea Institute of Science and Technology (KIST) \\
  \texttt{ \{hglee,hskim,drjay,y.choi\}@kist.re.kr} \\
}

\begin{document}

\maketitle

\begin{abstract}

Video Anomaly Detection (VAD) automatically identifies anomalous events from video, mitigating the need for human operators in large-scale surveillance deployments.
However, two fundamental obstacles hinder real-world adoption: domain dependency and real-time constraints---requiring near-instantaneous processing of incoming video.
To this end, we propose \ours{}, a zero-shot and real-time video anomaly detection paradigm.
Inspired by the human cognitive mechanism of instantly judging anomalies and reasoning in current scenes based on past experience, \ours{} operates in two stages: Recall and Respond.
In the \textit{offline recall} stage, an off-the-shelf LLM builds a pseudo-scene memory of both normal and anomalous captions without any reliance on real anomaly data.
In the \textit{online respond} stage, incoming video segments are embedded and matched against this memory via similarity search. By eliminating all LLM calls at inference time, Flashback delivers real-time VAD even on a consumer-grade GPU.
On two large datasets from real-world surveillance scenarios, UCF-Crime and XD-Violence, we achieve 87.3 AUC (+7.0 pp) and 75.1 AP (+13.1 pp), respectively, outperforming prior zero-shot VAD methods by large margins.

\end{abstract}

\section{Introduction}
\label{sec:intro}

Video Anomaly Detection (VAD) automatically identifies events that deviate from learned normal patterns in continuous video streams, overcoming the impracticality of manual monitoring in public safety~\cite{zhu2021video}, intelligent transportation~\cite{Bogdoll_2022_CVPR}, and industrial inspection systems~\cite{Roth_2022_CVPR}.
Currently, the number of surveillance cameras deployed worldwide is growing rapidly~\cite{businesswire2023surveillance,grandview2024video,Sultani_2018_CVPR},
generating volumes of video data that far exceed human monitoring capacity and making timely anomaly detection all but impossible.
Therefore, the practical adoption of automated VAD is urgently needed.

Real-world VAD deployment faces two fundamental obstacles: domain dependency and real-time constraints.
First, nearly all VAD paradigms---whether weakly-supervised~\cite{Sultani_2018_CVPR,ye2024vera}, one-class~\cite{Hasan_2016_CVPR,Lu_2013_ICCV}, or unsupervised~\cite{Thakare_2023_rareanom,tur2023diffusion}---require collecting and annotating domain-specific footage followed by model retraining for each new environment, imposing prohibitive time and cost burdens when target-domain samples are unavailable.
Second, since emergencies can occur at any time, processing each video segment should finish before the next one arrives.
Otherwise, in real-world scenarios with continuously arriving segments, some segments would have to be skipped to avoid latency accumulation, undermining uninterrupted anomaly detection.

\begin{figure}[t]
    \centering
    \includegraphics[width=\linewidth]{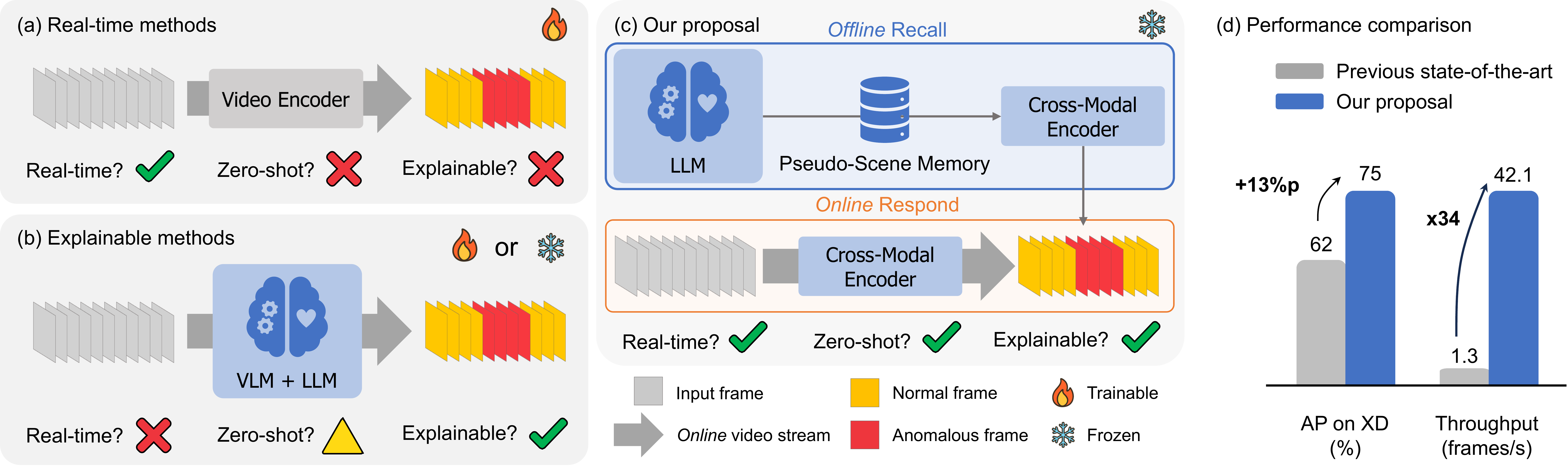}
    \figcaption{Bridging speed and reasoning}{
    (a) Real-time VAD keeps a light video encoder online but cannot work zero-shot or explain its decisions.
    (b) Explainable VAD adds a large VLM + LLM in the loop; reasoning is possible, yet speed drops and zero-shot ability is partial.
    (c) \ours{} moves the LLM offline, builds a pseudo-scene memory once, and uses a frozen cross-modal encoder at test time, so it is simultaneously real-time, zero-shot, and explainable.
    (d) On XD-Violence~\cite{Wu2020not} (XD), this design lifts AP by 13 percentage point and boosts throughput 34\(\times\) over the prior state-of-the-art.
    }
    \figcapmargin
    \label{fig:teaser}
\end{figure}

Recent research has addressed the challenges of domain generalization and real-time processing separately.
Zero-shot VAD techniques leverage pre-trained vision-language models~\cite{liu2023llava} (VLMs) and large language models~\cite{touvron2023llama2} (LLMs) to avoid any domain-specific retraining.
Among these, caption-and-score methods~\cite{gong2024anomalyclip,zanella2024harnessing} first generate segment captions via a VLM and then compute anomaly scores with an LLM,
but they suffer from the heavy computation of autoregressive captioning and noisy text outputs.
Prompt-based approaches~\cite{ahn2025anyanomaly,wu2024openvocabulary,ye2024vera} reduce LLM invocations by injecting optimized text prompts into the VLM inference stage, improving efficiency and domain transfer.
However, they often struggle to maintain coherent anomaly scores across temporally contiguous segments and remain sensitive to prompt vocabulary design.

Parallel efforts in real-time VAD aim to process each fixed-length segment before the next one arrives.
End-to-end weakly supervised models~\cite{karim2024reward} accelerate inference but still fall short of sub-second segment processing for unseen domains,
while density-estimation detectors~\cite{micorek2024mulde} achieve per-segment delays of around 200 ms
yet require domain-specific model updates to maintain accuracy.
Although zero-shot methods provide domain-agnostic adaptability and real-time ones deliver low latency, no existing approach unifies both capabilities.

In this paper, we propose the \ours{} paradigm to improve generalization and ensure consistently low-latency inference.
As illustrated in \figref{teaser}, \ours{} comprises two stages: offline pseudo-scene memory construction and online caption-retrieval inference.
In the offline stage, \ours{} leverages a frozen LLM~\cite{openai_gpt4o} to generate captions for a broad spectrum of normal and anomalous scenes without any video input, then converts each caption into an embedding via the video-text cross-modal encoder~\cite{bolya2025PerceptionEncoder,girdhar2023imagebind} and stores both captions and their embeddings in the memory.
In the online stage, incoming video is partitioned into fixed-length segments, whose embeddings are matched against the memory via similarity search to yield segment-level anomaly scores; these scores are then aggregated across segments and smoothed into frame-level predictions.

Despite this streamlined pipeline, we face two key challenges.
First, the encoder systematically biases representations toward anomalous captions even when processing normal content.
To address this, we introduce \textit{repulsive prompting}: wrapping normal and anomalous captions in distinct prompt templates so that their embeddings remain well separated.
Second, a residual skew toward anomalies can persist at inference time.
To correct this, we employ \textit{scaled anomaly penalization}, which attenuates anomalous-caption embedding magnitudes.

By fully decoupling the LLM from the online loop, \ours{} requires no additional data collection or fine-tuning.
Moreover, because anomaly detection finishes each fixed-length segment before the next one arrives—even when this interval is only one second—\ours{} enables genuine real-time VAD while providing human-readable explanations via the retrieved captions.
We evaluate \ours{} on two benchmark datasets—UCF-Crime~\cite{Sultani_2018_CVPR} and XD-Violence~\cite{Wu2020not}—and empirically demonstrate that it outperforms zero-shot, unsupervised, and one-class baselines.

The contributions of this work are as follows:

\textbf{A novel explainable zero-shot and real-time VAD paradigm}. We combine offline pseudo-scene memory construction using a frozen LLM with online caption retrieval inference to guarantee per-segment processing within its duration without any LLM calls at inference time.

\textbf{Novel techniques for mitigating anomaly bias}. We introduce \textit{repulsive prompting}, which applies distinct templates to normal and anomalous captions to prevent embedding collapse and enhance representation separation, and \textit{scaled anomaly penalization}, which attenuates anomalous-caption embedding magnitudes at inference time.

\textbf{Zero-shot pseudo-scene memory without real anomaly data}. Our pseudo-scene memory is built solely from LLM-generated captions—without using any actual anomalous video footage—yet still spans a broad range of normal and anomalous scenes.

\textbf{SOTA performance at real-time speed}. Even for segments as short as one second, our anomaly detector processes each in under one second—enabling genuine real-time VAD—and outperforms one-class and unsupervised methods on the UCF-Crime and XD-Violence.

\section{Related work}
\label{sec:related}

\textbf{Video anomaly detection.}
Early video anomaly detection (VAD) methods minimize reconstruction or other generative losses on normal-only or unlabeled footage~\cite{Hasan_2016_CVPR,Lu_2013_ICCV,Thakare_2023_rareanom,tur2023diffusion,tur2023exploring,Wang_2019_ICCV,Zaheer_2022_CVPR} and thus treat large reconstruction error as a signal of abnormality.
A parallel line formulates VAD as weakly-supervised multiple-instance learning, using video-level labels but no reliable frame labels~\cite{Sultani_2018_CVPR,zhang2019temporal,10.1109/TIP.2021.3062192}.
To broaden coverage, later work incorporates audio cues~\cite{Wu2020not,wu2025avadclip} or instruction-tunes detectors on a privately collected video-caption corpus~\cite{zhang2024holmes}.
More recently, researchers exploit the semantic priors of LLMs or VLMs: prompt tuning~\cite{ye2024vera} or lightweight adapters~\cite{zhang2024holmes} improve accuracy and even yield textual rationales.
However, all of these approaches still need target-domain videos or captions for fine-tuning; collecting and training on that data consumes both time and significant computational costs.
In contrast, \ours{} requires no additional data or gradient updates yet matches the accuracy of tuned systems while still providing explanations.

\textbf{Vision--language models for zero-shot VAD.}
Large language models~\cite{brown2020gpt3,ouyang2022instructgpt,wang2023selfinstruct,wei2022cot,wei2022flan} and vision-language models~\cite{alayrac2022flamingo,dai2023instructblip,li2023blip,peng2024kosmos2} already achieve strong few- or zero-shot accuracy on language benchmarks~\cite{clark2018arc,joshi2017triviaqa,kwiatkowski2019nq,paperno2016lambada,talmor2019commonsenseqa} and multimodal tasks such as visual question answering (VQA)~\cite{VQA,VQAv2,GQA,OKVQA}.
LAVAD~\cite{zanella2024harnessing} extends this power to anomaly detection: it captions every video segment with a VLM and scores each caption with an LLM, achieving competitive zero-shot performance and human-readable explanations.
Subsequent work refines prompts~\cite{ye2024vera}, adds graph modules for modeling temporal structure~\cite{shao2025eventvad}, or fuses audio via audio-language models~\cite{prasad2024mca}.
Nevertheless, all of these pipelines keep an auto-regressive model within the repetitive operation, so inference slows to roughly around one frame per second, so latency varies with output caption length.
On the contrary, \ours{} leverages the knowledge base of an LLM~\cite{openai_gpt4o} \emph{offline}: we generate a pseudo-scene memory without any visual input, store it once, and then perform real-time retrieval with a video-text encoder.
At inference time, we only look up the most similar caption for each segment, assign its anomaly label, and use the caption itself as an immediate textual explanation, sustaining 42.06 fps on a single commercial GPU while avoiding any online LLM calls.

\section{The \ours{} paradigm}
\label{sec:method}

Our framework, \ours{}, casts video anomaly detection as a two‐stage, memory‐driven retrieval task.
Drawing on how humans instantly detect and explain anomalies by recalling past experiences~\cite{Bar2007,Friston2005,Summerfield2014},
\ours{} first enters a recall phase in which a single LLM invocation builds diverse captions of normal/anomalous scenarios.
In the offline recall stage, an LLM builds a pseudo-scene memory encompassing both scenarios.
In the online respond stage, incoming video segments are converted into embeddings and assessed for anomalies via similarity search against that memory.
This design enables \ours{} to deliver fast, accurate zero-shot VAD without any LLM calls at inference time.

\begin{figure}[t]
    \centering
    \includegraphics[width=\linewidth]{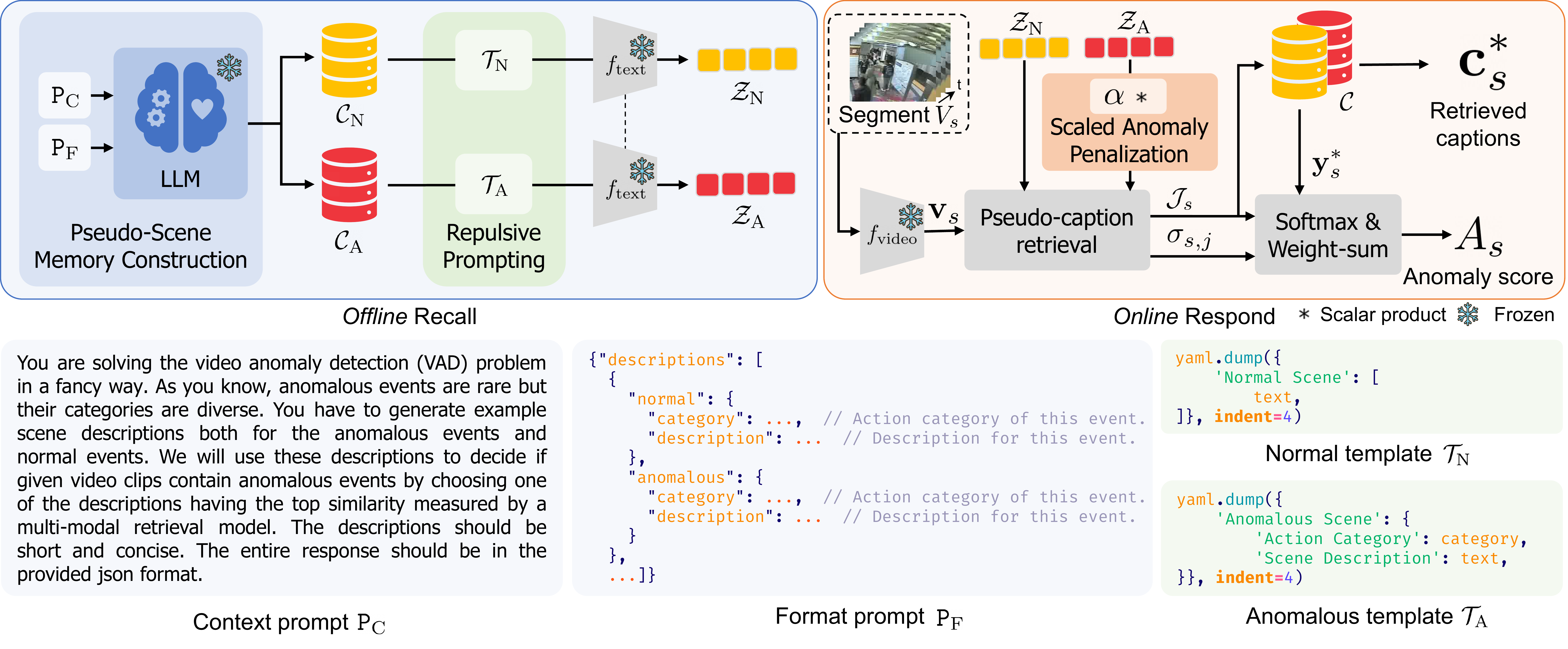}
    \figcaption{Overview of \ours{}}{
    \ours{} operates in two disjoint stages.
    \textbf{Offline Recall}: a frozen LLM~\cite{openai_gpt4o} generates a diverse set of normal and anomalous scene sentences using context and format prompts \(\mathtt{P}_\text{C},\mathtt{P}_\text{F}\), which are embedded by a frozen video-text encoder and stored in a million-entry \textit{Pseudo-Scene Memory} \(\mathcal{C}_\text{N}, \mathcal{C}_\text{A}\).
    \textit{Repulsive Prompting} widens the separation between normal and anomalous embeddings, countering the encoder's bias.
    \textbf{Online Respond}: we embed each incoming segment \(V_s\), retrieve its top-\(K\) matches from the memory, and debias the resulting similarities with \textit{Scaled Anomaly Penalization}.
    The resulting scores, together with the retrieved sentences, provide real-time anomaly alerts and concise textual rationales.
    }
    \figcapmargin
    \label{fig:method}
\end{figure}

\vspace{-0.3em}
\subsection{Problem statement: zero-shot VAD}
\label{sec:problem}

Let a video be a frame sequence \(V=(I_t)_{t=1}^{T}\).
The task of video anomaly detection (VAD) is to assign an anomaly score \(s_t\in[0,1]\) to every frame \(I_t\).
Prior work differs mainly in how much supervision each setting uses:
i) weakly-supervised methods train on videos that carry only a video-level label (anomalous or normal) but no frame labels;
ii) one-class methods see only normal videos during training and detect any deviation at inference time;
and iii) unsupervised methods assume no labels at all.
In zero-shot VAD, we go one step further: the detector receives \emph{no} target-domain videos or labels of any kind during training, yet must still output frame-wise scores on unseen data.
Formally, our training set is empty (\(\mathcal{V}_{\mathrm{train}}=\varnothing\)).

\textbf{VAD as pseudo-caption retrieval.}
We cast video anomaly detection as an \textit{online} retrieval task on an \textit{offline} pseudo-scene memory.
Given a segment \(V_s\), a frozen video-text encoder produces a feature \(\mathbf{v}_s\) and compares it with pre-encoded caption vectors \(\mathbf{t}_j\).
The top \(K\) matches define soft weights \(w_{s,k}\);
the segment score is a label average \(A_s=\sum_k w_{s,k}y_{j'_k}\),
and the matched captions serve as explanations.
This needs one encoder pass and a few dot products, so inference runs in real time.
Splitting long videos into overlapping segments and smoothing the scores yields frame-level predictions \(p_t\).

\textbf{Two challenges.}
First, the pseudo-captions must span a broad range of scenes.
We generate millions of normal and anomalous pseudo-captions with a single LLM prompt, then show in \secref{abl} that the coverage is sufficient.
Second, raw captions lie too close in the embedding space and show poor performance (AUC 75.13 on UCF-Crime).
We address this in two ways.
\emph{Repulsive prompting} adds a class keyword and a lightweight wrapper, pushing normal and anomalous text features apart without any training.
\emph{Scaled anomaly penalization} further reduces the influence of anomalous captions by down-weighting their similarity scores.

\vspace{-0.5em}
\subsection{Offline recall (data preparation)}
\label{sec:offline}

\textbf{Pseudo-caption memory.}
This step builds the offline memory queried at inference time.
We run the LLM~\cite{openai_gpt4o} with two prompts.
\textit{Context prompt} \(\mathtt{P}_{\text{C}}\):
The model is told to act as a VAD assistant.
It must produce short captions for both normal and anomalous events. %, noting that anomalies are rare but diverse.
\(\mathtt{P}_{\text{C}}\) also explains that the captions will later be ranked by a cross-modal encoder, so they should be concise and informative.
\textit{Format prompt} \(\mathtt{P}_{\text{F}}\):
To keep the output machine-parsable, we supply a schema with two
fields, \texttt{"normal"} and \texttt{"anomalous"},
each holding an \texttt{"action category"} and a free-form
\texttt{"description"}.

The LLM returns ordered lists of captions
\(\mathcal{C}_{\text{N}}=(c^{\text{N}}_{1},\dots,c^{\text{N}}_{N_{\text{N}}})\) (normal) and \(\mathcal{C}_{\text{A}}=(c^{\text{A}}_{1},\dots,c^{\text{A}}_{N_{\text{A}}})\)
(anomalous), together with their category lists
\(\mathcal{K}_{\text{N}}=(\kappa^{\text{N}}_{1},\dots,\kappa^{\text{N}}_{N_{\text{N}}})\) and \(\mathcal{K}_{\text{A}}=(\kappa^{\text{A}}_{1},\dots,\kappa^{\text{A}}_{N_{\text{A}}})\).
We concatenate them to form
\begin{equation}
\mathcal{C}= \mathcal{C}_{\text{N}} \oplus \mathcal{C}_{\text{A}},\quad
\mathcal{K}= \mathcal{K}_{\text{N}} \oplus \mathcal{K}_{\text{A}},\quad
Y=(\underbrace{0,\dots,0}_{N_{\text{N}}},\underbrace{1,\dots,1}_{N_{\text{A}}}),
\end{equation}
where \(Y\) stores the binary anomaly flags.
All captions are encoded once with the text branch \(f_\text{text}\)
and cached, so no further LLM calls or fine-tuning are required during
inference.

\textbf{Repulsive prompting.}
Pseudo-captions that describe similar scenes often fall close together in the embedding space even when one is normal and the other is anomalous.
We introduce a simple yet effective idea, repulsive prompting, to push the two groups apart without altering their core meaning.
We insert one word within every caption.
The word is \texttt{"Normal"} for routine events and \texttt{"Anomalous"} for abnormal events.
Variants such as \texttt{"Abnormal"} or \texttt{"Anomaly"} were tested, but \texttt{"Anomalous"} gives the clearest separation, so we use it throughout.
Second, each caption then passes through a template.
We call the combined keyword-plus-wrapper \(\mathcal{T}_\text{N}\) for normal and \(\mathcal{T}_\text{A}\) for anomalous.
We note that \(\mathcal{T}_\text{N} \neq \mathcal{T}_\text{A}\).
Any lightweight data-formatting wrapper works
and can embed an action category while leaving the main sentence intact.
Encoding the templated captions yields two caption feature sets
\(\mathcal{Z}_\text{N}=f_{\text{text}}(\mathcal{T}_\text{N}(\mathcal{C}_\text{N}, \mathcal{K}_\text{N}))\),
\(\mathcal{Z}_\text{A}=f_{\text{text}}(\mathcal{T}_\text{A}(\mathcal{C}_\text{A}, \mathcal{K}_\text{A}))\).
Both \(\mathcal{Z}_\text{N}\) and \(\mathcal{Z}_\text{A}\) lie in \(\mathbb{R}^D\).
The angle between the centroids of \(\mathcal{Z}_\text{N}\) and \(\mathcal{Z}_\text{A}\) is larger than the angle obtained with raw captions, and \secref{abl} shows that this wider separation leads to fewer false positives.
It's worth noting that the whole step adds only a few tokens and needs no training.

\vspace{-0.3em}
\subsection{Online respond (inference)}
\label{sec:online}

\textbf{Scaled anomaly penalization.}
We observe that caption vectors for anomalous events tend to form smaller angles with video features than do normal captions.
Therefore, shrinking those vectors before dot product computation is more effective than clipping the score after retrieval.
To damp the inherent bias towards anomalous captions, we rescale their embeddings before retrieval:
for every \(\mathbf{t}_j \in \mathcal{Z}_{\text{A}}\), we set
\(\tilde{\mathbf{t}}_j = \alpha \,\mathbf{t}_j\)
with \(\alpha \in (0,1)\).
The factor \(\alpha\) lowers the magnitude of dot products for anomalous captions,
reducing spurious matches with adding little computational overhead.

\textbf{Pseudo-caption retrieval.}
We chop a test video into segments \( ( V_s ) \) of \( T_\text{segment}\) seconds with overlap \(T_\text{overlap} < T_\text{segment}\) seconds.
We obtain features of \(s\)-th segment \( \mathbf{v}_s = f_\text{video} (V_s) \).
For every caption embedding \(\mathbf{t}_j \in \mathcal{Z} \), we compute the dot products \(\sigma_{s,j}=\mathbf{v}_s^\top\mathbf{t}_j\).
We keep the set of indices of the \(K\) largest products \(\mathcal{J}_s\) % \( \mathcal{J}_s = ( j_1^\prime, \dots, j_K^\prime )_s \)
and convert their products to weights \( (w_{s,k})_{k=1}^K \) by applying softmax through \(k\). % \(w_{s,k}=\mathtt{softmax}_k(\sigma_{s,j_k})\).
Then, we obtain the segment anomaly score \( A_s \) as the weighted average of retrieved anomaly flags \(\mathbf{y}^*_s\).
The retrieved captions \( \mathbf{c}^*_s \) are returned as instant textual explanations of the segment.

\textbf{Frame-level score refinement.}
Let \(\mathcal{S}_t\) be the set of indices of segments that cover frame \(t\).
We average their scores \(p_t=\tfrac{1}{|\mathcal{S}_t|}\sum_{s\in \mathcal{S}_t} A_s\) and convolve the resulting sequence with a 1-D Gaussian.
The smoothed curve \((p_t)\) is our final frame-wise prediction.

\vspace{-0.3em}
\subsection{Computational complexity}
\label{sec:comp}

In our real-time setting, we compare the online per-segment computational cost of our retrieval-driven approach against an existing VLM-based pipeline by decomposing each into video encoding and text processing components.
Let \(C_{\text{video}}\) denote the cost of extracting visual features (common to both methods), and define
\(
C_{\text{VLM}}
= C_{\text{video}}
+ C_{\text{LLM}}
\),
\(
C_\text{\ours{}}
= C_{\text{video}}
+ C_{\text{retrieve}}.\)
We therefore need only compare \(C_{\text{LLM}}\) to \(C_{\text{retrieve}}\).

The LLM cost can be treated as a black box with quadratic dependence on the total sequence length. Setting
\(
L = L_{\text{in}} + L_{\text{out}}
= T \, L_{\text{image}} + L_{\text{prompt}} + L_{\text{out}}
\),
and \(c_{\text{LLM}}\) as work to process a single token through all \(M_{\text{LLM}}\) layers of size \(D_{\text{LLM}}\),
we have
\(
  C_{\text{LLM}}
  = \mathcal{O}\bigl(L^2 \, c_{\text{LLM}}\bigr)
  \approx \mathcal{O}\bigl(L^2 \, D_{\text{LLM}}^2 \, M_{\text{LLM}}\bigr).
\)
By contrast, our retrieval cost scales linearly in the number of stored caption vectors \(N\) and their dimension \(D\) (normally in the thousands):
\(
  C_{\text{retrieve}}
  = \mathcal{O}\bigl(N\,D + N \log N\bigr)
  \approx \mathcal{O}(N\,D).
\)

Because in practical settings, \(L\) is large (\eg dozens of \(N\)) and \(D_{\text{LLM}}^2 M_{\text{LLM}} \gg D\), it follows that
\begin{equation}
  C_{\text{VLM}}
  \approx C_{\text{video}}
  + \mathcal{O}\bigl(L^2 \, D_{\text{LLM}}^2 \, M_{\text{LLM}}\bigr)
  \;\gg\;
  C_{\text{video}}
  + \mathcal{O}(N\,D)
  \approx C_\text{\ours{}}.
\end{equation}
We note that \(C_\text{LLM} \gg C_\text{video} \gg C_\text{retrieve}\).
Though \(N\) may reach millions, replacing the LLM call with a retrieval step reduces the per-segment computational complexity by several orders of magnitude.

\section{Experimental results}
\label{sec:result}

We structure our evaluation to validate all core properties of \ours{}—zero-shot SOTA accuracy without any target-domain training (\secref{sota}), real-time performance (\secref{through}), and explainability via retrieved captions and event categories (\secref{qual_main} \& \secref{abl}).
We also examine anomaly-score correlation with true anomalousness (\secref{qual_main}), and assess how repulsive prompting (RP) and scaled anomaly penalization (SAP) reduce false positives, as well as SAP's sensitivity to the scale factor \(\alpha\) (\secref{abl}).

\vspace{-0.5em}
\subsection{Experimental setup}
\label{sec:setup}

\textbf{Datasets.}
We evaluate on two large-scale benchmarks.
\textbf{UCF-Crime}~\cite{Sultani_2018_CVPR} (UCF) contains 290 test videos (140 abnormal) with 13 anomaly types.
\textbf{XD-Violence}~\cite{Wu2020not} (XD) provides 800 test videos (500 abnormal) covering six categories.

\textbf{Metrics.}
Following prior work~\cite{ye2024vera,zanella2024harnessing} we report the
area under the frame-level ROC curve (AUC) for both datasets and the
frame-level average precision (AP) for XD-Violence.

\textbf{Baselines.}
We compare with representative methods at each supervision level:
weakly-supervised
\cite{10.1609/aaai.v37i1.25112,joo2023cliptsa,Li_Liu_Jiao_2022,Sultani_2018_CVPR,Tian_2021_ICCV,WuHCFL22,10.1109/TIP.2021.3062192,Wu2020not,wu2023vadclip,ye2024vera,Zaheer_2022_CVPR,zhang2024holmes},
one-class
\cite{Hasan_2016_CVPR,Lu_2013_ICCV,Wang_2019_ICCV},
unsupervised
\cite{Thakare_2023_rareanom,Thakare_2023_WACV,tur2023diffusion,tur2023exploring,Zaheer_2022_CVPR},
and zero-shot
\cite{zanella2024harnessing}.
For Holmes-VAD~\cite{zhang2024holmes}, we quote the numbers reported in VERA~\cite{ye2024vera} because the fine-grained annotations for instruction tuning are not public.

\textbf{Implementation.}
The pseudo-caption memory is generated once with \texttt{gpt-4o-2024-08-06}~\cite{openai_gpt4o}, costing \$181.43 and 76 hours for one million normal-anomalous pairs.
The frozen cross-modal encoder is ImageBind~\cite{girdhar2023imagebind} (\ours{}-IB) or PerceptionEncoder~\cite{bolya2025PerceptionEncoder} (\ours{}-PE).
Unless noted, we use \(T_{\text{segment}}=1\)s,\, \(T_{\text{overlap}}=0\)s,\, \(T_{\text{sample}}=16\) frames,\, input resolution \(448{\times}336\),
Gaussian kernel with 100-frame width and \(\sigma=0.5\).
Lastly, we set the number of captions for retrieval \(K\) as 10.
All experiments run on a single RTX 3090.

\vspace{-0.3em}
\subsection{Comparison with state-of-the-art methods}
\label{sec:sota}

\begin{table}[t]
\centering
\caption{
    \textbf{Comparison with state-of-the-art video anomaly detectors on UCF-Crime~\cite{Sultani_2018_CVPR} and XD-Violence~\cite{Wu2020not}.}
    Methods are grouped by supervision level (weakly-supervised, one-class, unsupervised, and zero-shot).
    \ours{} attains the highest accuracy on both datasets and is the first approach that is simultaneously zero-shot, real-time, and explainable.
    \textbf{Bold} numbers mark the top result.
    Scores marked \textsuperscript{*} are reported by CLIP-TSA~\cite{joo2023cliptsa}, and
    scores marked \textsuperscript{\textdagger} are reported by VERA~\cite{ye2024vera}.
}
\tabcapmargin
\resizebox{0.6\columnwidth}{!}{
\begin{tabular}{l cc c cc}
\toprule
\multirow{2}[2]{*}{Method}
& \multirow{2}[2]{*}{Explainable?}
& \multirow{2}[2]{*}{Real-time?}
& UCF-Crime & \multicolumn{2}{c}{XD-Violence}
\\
\cmidrule(lr){4-4}
\cmidrule(lr){5-6}
&  &  & AUC (\%) & AP (\%) & AUC (\%)
\\
\midrule
{\textit{Weakly-Supervised}} &&&&&
\\
{Sultani \etal~\cite{Sultani_2018_CVPR}}
& \phantom{\xmark} & \cmark
& 77.92 & - & -
\\
GCL~\cite{Zaheer_2022_CVPR}
& \phantom{\xmark} & \cmark
& 79.84 & - & -
\\
Wu \etal~\cite{Wu2020not}
& \phantom{\xmark} & \cmark
& 82.44 & 73.20 & -
\\
RTFM~\cite{Tian_2021_ICCV}
& \phantom{\xmark} & \cmark
& 84.03 & 77.81 & -
\\
Wu \& Liu~\cite{10.1109/TIP.2021.3062192}
& \phantom{\xmark} & \cmark
& 84.89 & 75.90 & -
\\
MSL~\cite{Li_Liu_Jiao_2022}
& \phantom{\xmark} & \cmark
& 85.62 & 78.58 & -
\\
S3R~\cite{WuHCFL22}
& \phantom{\xmark} & \cmark
& 85.99 & 80.26 & -
\\
MGFN~\cite{10.1609/aaai.v37i1.25112}
& \phantom{\xmark} & \cmark
& 86.98 & 80.11 & -
\\
CLIP-TSA~\cite{joo2023cliptsa}
& \phantom{\xmark} & \cmark
& 87.58 & 82.17\makebox[0pt][l]{\textsuperscript{*}} & -
\\
VadCLIP~\cite{wu2023vadclip}
& \phantom{\xmark} & \cmark
& 88.02 & 84.51 & -
\\
Holmes-VAD~\cite{zhang2024holmes}
& \cmark & \phantom{\xmark}
& 84.61\makebox[0pt][l]{\textsuperscript{\textdagger}} & 84.96\makebox[0pt][l]{\textsuperscript{\textdagger}} & -
\\
VERA~\cite{ye2024vera}
& \cmark & \phantom{\xmark}
& 86.55 & 70.54 & 88.26
\\
\midrule
\textit{One-Class} &&&&&
\\
Hasan \etal~\cite{Hasan_2016_CVPR}
& \phantom{\xmark} & \cmark
& - & - & 50.32
\\
Lu \etal~\cite{Lu_2013_ICCV}
& \phantom{\xmark} & \cmark
& - & - & 53.56
\\
BODS~\cite{Wang_2019_ICCV}
& \phantom{\xmark} & \cmark
& 68.26 & - & 57.32
\\
GODS~\cite{Wang_2019_ICCV}
& \phantom{\xmark} & \cmark
& 70.46 & - & 61.56
\\
\midrule
\textit{Unsupervised} &&&&&
\\
GCL~\cite{Zaheer_2022_CVPR}
& \phantom{\xmark} & \cmark
& 74.20 & - & -
\\
Tur \etal~\cite{tur2023exploring}
& \phantom{\xmark} & \cmark
& 65.22 & - & -
\\
Tur \etal~\cite{tur2023diffusion}
& \phantom{\xmark} & \cmark
& 66.85 & - & -
\\
DyAnNet~\cite{Thakare_2023_WACV}
& \phantom{\xmark} & \cmark
& 79.76 & - & -
\\
RareAnom~\cite{Thakare_2023_rareanom}
& \phantom{\xmark} & \cmark
& - & - & 68.33
\\
\midrule
\textit{Zero-shot} &&&&&
\\
LAVAD~\cite{zanella2024harnessing}
& \cmark & \phantom{\xmark}
& 80.28 & 62.01 & 85.36
\\
\textbf{\ours{}-IB (Ours)}
& \cmark & \cmark
& 81.65 & 60.13 & 83.52
\\
\textbf{\ours{}-PE (Ours)}
& \cmark & \cmark
& \textbf{87.29} & \textbf{75.13} & \textbf{90.54}
\\
\bottomrule
\end{tabular}
}
\vspace{-.5em}
\label{tab:sota}
\end{table}

\ours{} achieves the best results in every zero-shot setting, as summarised in \tabref{sota}.
\ours{} raises the state of the art by a large gain of XD AP over zero-shot LLaVA-1.5~\cite{zanella2024harnessing,liu2023improvedllava} and LAVAD~\cite{zanella2024harnessing} (50.26, 62.01 \vs 75.13).
The proposed method (XD AUC 90.54) also surpasses strong unsupervised (\vs 68.33), one-class (\vs 61.56), and several weakly-supervised baselines.
Particularly, it exceeds the explainable VERA~\cite{ye2024vera} on XD AP (70.54 \vs 75.13).
To our knowledge, \ours{} is the first VAD system that is simultaneously zero-shot, real-time, and able to return human-readable explanations.

\vspace{-0.3em}
\subsection{Throughput evaluation protocol}
\label{sec:through}

A detector processes the video in fixed segments of length \(T_{\text{segment}}\).
Consecutive segments overlap by \(T_{\text{overlap}}\),
so the time budget before the next segment arrives is \(T_{\text{decision}} = T_{\text{segment}} - T_{\text{overlap}}\).
Let \(T_{\text{process}}\) be the wall-clock time required to analyze one segment.
We call a detector \emph{real-time} when
\begin{equation}
T_{\text{decision}} \le 1\ \text{second} \qquad\text{and}\qquad T_{\text{process}} \le T_{\text{decision}}.
\label{eq:real}
\end{equation}
In our experiments, \(T_{\text{segment}} = 1\text{s}\) and \(T_{\text{overlap}} = 0\text{s}\), so the decision period equals one second. \ours{} completes a segment in \(T_{\text{process}} = 0.713\text{s}\), easily meeting the requirement. LAVAD~\cite{zanella2024harnessing} relies on sequential VLM calls and a non-causal caption-refinement step, so its decision period is unbounded (\(T_\text{decision} = \infty\)) and real-time response cannot be guaranteed.

\textbf{Throughput and accuracy.}
\tabref{lat} lists frame rate and test accuracy.
On XD-Violence~\cite{Wu2020not}, \ours{} improves AP from LAVAD's 62.01 to 75.13, while boosting frame rate from 1.26 fps to 42.06 fps, a speed-up of roughly thirty-four times.

\vspace{-0.3em}
\subsection{Qualitative analysis}
\label{sec:qual_main}

\begin{figure}[t]
    \centering
    \includegraphics[width=0.99\linewidth]{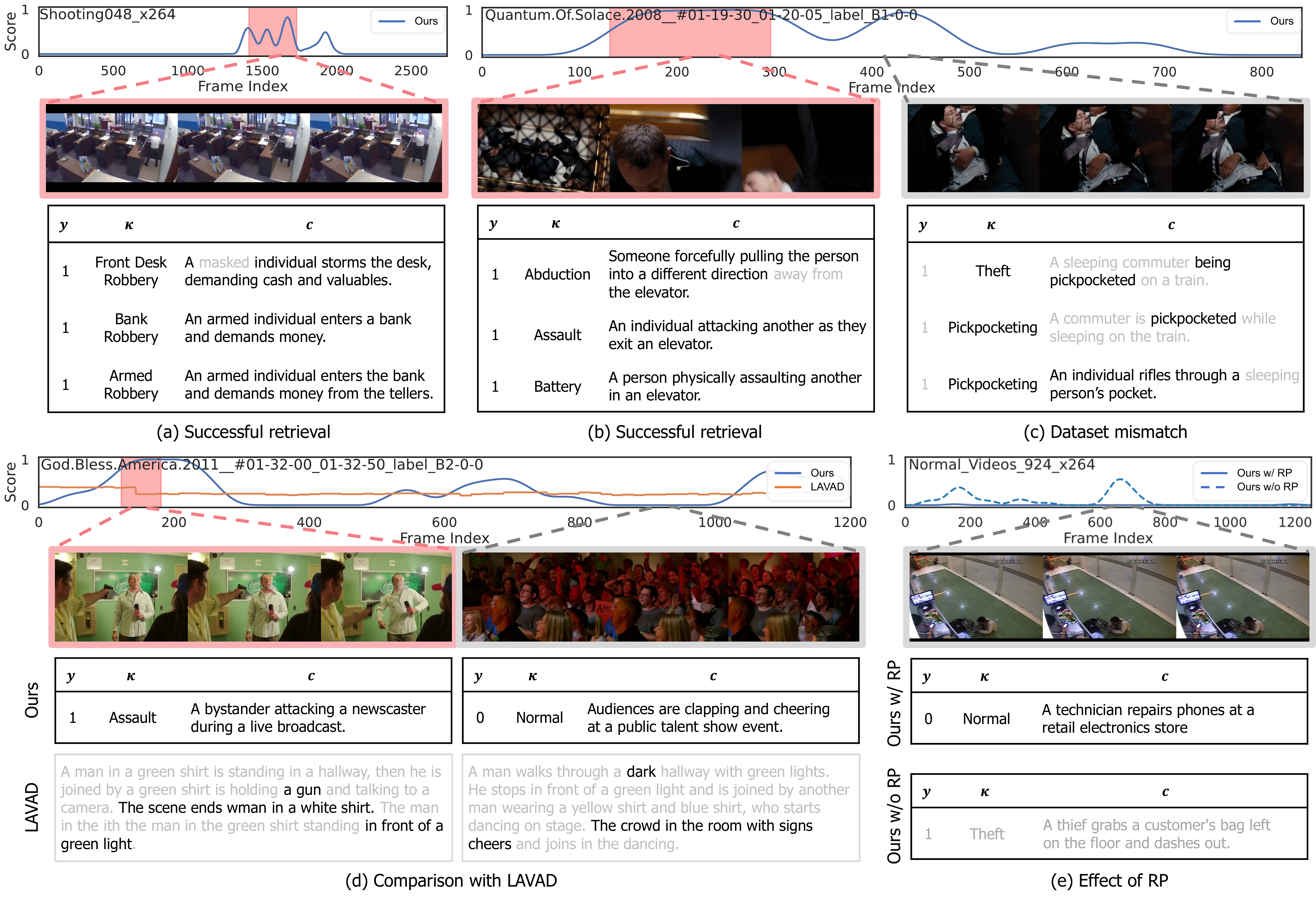}
    \figcaption{Qualitative examples}{
    The plots show frame-wise anomaly curves.
    Red boxes on both the video strip and the plot mark \colorbox{myred}{\strut \;ground-truth anomalous intervals\;}.
    For selected frames we list the retrieved category-caption pairs \((\kappa,c)\) and their anomaly flags \(y\).
    Black text denotes a correct description, \textcolor{Gray}{gray} text an incorrect one.
    (a) \& (b) The top captions describes the event precisely.\;
    (c) \ours{} flags ``Pickpocketing'' as abnormal, but XD-Violence~\cite{Wu2020not} treats it as normal.\;
    (d) LAVAD~\cite{zanella2024harnessing} misses short anomalies and often outputs malformed sentences, whereas \ours{} detects the event and returns a concise caption.\;
    (e) Removing repulsive prompting (RP) causes frequent false alarms on a normal clip. %; adding RP suppresses them and retrieves a relevant caption.
    }
    \figcapmargin
    \vspace{-.5em}
    \label{fig:qual_main}
\end{figure}

\figref{qual_main} presents qualitative results that shed further light on \ours{}.
\figref{qual_main} (a) and (b) display the per-frame anomaly-score curve at the top, and for each segment lists the \(K\) retrieved captions \(c\) together with their anomaly flags \(y\) and categories \(\kappa\).
The examples in (a) and (b) are striking for three reasons:
i) nearly exact pseudo-captions—generated with no video input—already exist in the memory;
ii) the retrieval step picks those captions as top matches;
and iii) the anomaly flag and the score reflect the level of anomalousness in the scene.
(c) shows a failure that is also instructive.
The detector flags ``Pickpocketing'' as abnormal because our memory marks that activity as an anomaly,
yet XD-Violence treats it as normal.
Although the prediction counts as a false positive under the benchmark,
the explanation text is still plausible.

We compare score curves from LAVAD~\cite{zanella2024harnessing} and \ours{} in (d).
LAVAD produces flat curves, so short anomalies are easy to miss.
In contrast, \ours{} shows clear slopes where the event happens, making threshold selection much easier for downstream users.
The captions are also concise and free of garbled sentences often produced by large-model captioning~\cite{li2024multi}.
Removing repulsive prompting (RP) increases false positives markedly, especially on static normal shots as illustrated in (e).

\vspace{-.5em}
\subsection{Ablation study}
\label{sec:abl}

\begin{table}[t]
\centering
\vspace{-1.8em}
\caption{
    \textbf{Real-time performance comparison:} We compare \ours{} with LAVAD~\cite{zanella2024harnessing} in terms of decision period \(T_{\mathrm{decision}}\), processing time \(T_{\mathrm{process}}\), and frame rate. Cases meeting the real-time requirement \eqnref{real} are highlighted in \textcolor{Green}{green}, and those that do not are highlighted in \textcolor{red}{red}. \ours{} achieves approximately 34\(\times\) higher throughput, shorter processing times, and higher accuracy com\-pared to LAVAD, thereby satisfying the real-time criterion. Best values are shown in \textbf{bold}.
}
\tabcapmargin
\mpage{0.7}{
\resizebox{\columnwidth}{!}{
\begin{tabular}{l c cc cc cc}
\toprule
\multirow{2}[2]{*}{Method}
& UCF-Crime
& \multicolumn{2}{c}{XD-Violence}
& \multirow{2}[2]{*}{\shortstack[c]{\(T_\text{decision}\vphantom{_p}\)\\ (sec)}}
& \multirow{2}[2]{*}{\shortstack[c]{\(T_\text{process}\vphantom{_p}\)\\ (sec)}}
& \multirow{2}[2]{*}{\shortstack[c]{Frame rate\vphantom{p}\\ (frames/sec)}}
& \multirow{2}[2]{*}{\shortstack[c]{Speed up\\ (times)}}
\\
\cmidrule(lr){2-2}
\cmidrule(lr){3-4}
& AUC (\%) & AP (\%) & AUC (\%)
\\
\midrule
LAVAD~\cite{zanella2024harnessing}
& 80.28 & 62.01 & 85.36
& {\color{red} \(\infty\)}
& {\color{Green} 23.810}
& \phantom{0}1.26
& \phantom{0}\(1.0 \times \)
\\
\ours{} (Ours)
& \textbf{87.29} & \textbf{75.13} & \textbf{90.54}
& {\color{Green} \textbf{1.0}}
& \phantom{2}{\color{Green} \textbf{0.713}}
& \textbf{42.06}
& \(\textbf{33.9} \times \)
\\
\bottomrule
\end{tabular}
}
\vspace{-0.6em}
}
\label{tab:lat}
\end{table}

\begin{table*}[t]
\centering
\caption{\textbf{Ablation study.}
We report AUC on UCF-Crime and both AUC and AP on XD-Violence.
(a) We randomly sample four disjoint subsets of 10k pseudo-captions from the 1M pseudo-scene memory (time and budget constraints prevent larger sweeps).
(b) We additionally list the angle \(\theta\) (in degrees) between the normal and anomalous centroids.
(d) \& (e) Frame rate (fps) is provided alongside accuracy.
(e) Decision period \(T_{\text{decision}}\) and processing time \(T_{\text{process}}\) are colored \textcolor{Green}{green} when they satisfy the real-time condition \eqnref{real}
and \textcolor{red}{red} otherwise.
The \textbf{best} value in each column is bold; the configuration adopted for the main system is \colorbox{mygray}{\strut \;gray-shaded\;}.
}

\mpage{0.29}{\fontsize{8}{10}\selectfont (a) Stability and reproducibility of memory.}\hfill
\mpage{0.39}{\fontsize{8}{10}\selectfont (b) Effectiveness of repulsive prompting.}\hfill
\mpage{0.25}{\fontsize{8}{10}\selectfont (c) Impact of the number of retrieved captions \(K\).}\hfill
\\
\mpage{0.285}{
\resizebox{\columnwidth}{!}{
\begin{tabular}{c c cc}
\toprule
\multirow{2}[2]{*}{Seed} & UCF-Crime & \multicolumn{2}{c}{XD-Violence}
\\
\cmidrule(lr){2-2}
\cmidrule(lr){3-4}
& AUC (\%) & AP (\%) & AUC (\%)
\\
\midrule
A
& 84.75 & 74.57 & 90.21
\\
B
& 84.96 & 74.12 & 90.05
\\
C
& 84.21 & 74.50 & 90.16
\\
D
& 83.61 & 74.10 & 90.10
\\
\midrule
Overall
& 84.38\makebox[0pt][l]{\small \(_{\pm 0.60}\)} & 74.32\makebox[0pt][l]{\small \(_{\pm 0.25}\)} & 90.13\makebox[0pt][l]{\small \(_{\pm 0.07}\)}
\\
\bottomrule
\end{tabular}
}
}
\hfill
\mpage{0.39}{
\resizebox{\columnwidth}{!}{
\begin{tabular}{c c ccc}
\toprule
\multirow{2}[1]{*}{Strategy} & UCF-Crime & \multicolumn{2}{c}{XD-Violence}
& \multirow{2}[2]{*}{\shortstack[c]{\(\theta\) (\(\uparrow\))\\ (deg)}}
\\
\cmidrule(lr){2-2}
\cmidrule(lr){3-4}
& AUC (\%) & AP (\%) & AUC (\%)
\\
\midrule
\xmark
& 74.98 & 71.01 & 87.08
& \phantom{0}8.12
\\
Lin. alg. op.
& 81.56 & 64.98 & 83.04
& \phantom{0}8.12
\\
RP (keyword-only)
& 81.24 & 72.20 & 88.42
& 27.79
\\
RP (template-only)
& 82.12 & 72.21 & 88.82
& 23.49
\\
\rowcolor{mygray}
RP
& \textbf{87.29} & \textbf{75.13} & \textbf{90.54}
& \textbf{33.29}
\\
\bottomrule
\end{tabular}
}
}
\hfill
\mpage{0.25}{
\resizebox{\columnwidth}{!}{
\begin{tabular}{c c cc}
\toprule
\multirow{2}[2]{*}{\(K\)} & UCF-Crime & \multicolumn{2}{c}{XD-Violence}
\\
\cmidrule(lr){2-2}
\cmidrule(lr){3-4}
& AUC (\%) & AP (\%) & AUC (\%)
\\
\midrule
1
& 82.00 & 73.55 & 88.46
\\
5
& 85.66 & 74.84 & 90.19
\\
\rowcolor{mygray}
10
& \textbf{87.29} & \textbf{75.13} & \textbf{90.54}
\\
20
& 86.84 & 75.08 & 90.71
\\
40
& 86.31 & 74.86 & 90.73
\\
\bottomrule
\end{tabular}
}
}
\hfill

\mpage{0.43}{\fontsize{8}{10}\selectfont (d) Effectiveness and efficiency of the size of memory.}
\mpage{0.5}{\fontsize{8}{10}\selectfont (e) Effectiveness and efficiency of video segment parameters.}
\\
\mpage{0.03}{\ }
\mpage{0.35}{
\resizebox{\columnwidth}{!}{
\begin{tabular}{c c c cc}
\toprule
\multirow{2}[2]{*}{\shortstack[c]{\# Caption\\ pairs}}
& \multirow{2}[2]{*}{\shortstack[c]{Frame rate\\ (frames/s)}}
& UCF-Crime & \multicolumn{2}{c}{XD-Violence}
\\
\cmidrule(lr){3-3}
\cmidrule(lr){4-5}
&
& AUC (\%) & AP (\%) & AUC (\%)
\\
\midrule
\phantom{0,0}10,000
& \textbf{42.95}
& 82.61 & 72.17 & 88.63
\\
\phantom{0,0}50,000
& 42.95
& 84.74 & 74.30 & 90.01
\\
\phantom{0,}100,000
& 42.95
& 84.38 & 74.32 & 90.13
\\
\phantom{0,}500,000
& 42.58
& 85.20 & 75.11 & 90.52
\\
\rowcolor{mygray}
1,000,000
& 42.06
& \textbf{87.29} & \textbf{75.13} & \textbf{90.54}
\\
\bottomrule
\end{tabular}
}
}
\mpage{0.04}{\ }
\mpage{0.5}{
\resizebox{\columnwidth}{!}{
\begin{tabular}{ccc ccc ccc}
\toprule
\multirow{2}[2]{*}{\shortstack[c]{\(T_\text{segment}\vphantom{_p}\)\\ (sec)}}
& \multirow{2}[2]{*}{\shortstack[c]{\(T_\text{stride}\vphantom{_p}\)\\ (sec)}}
& \multirow{2}[2]{*}{\shortstack[c]{\(T_\text{sample}\)\\ (frames)}}
& \multirow{2}[2]{*}{\shortstack[c]{\(T_\text{decision}\vphantom{_p}\)\\ (sec)}}
& \multirow{2}[2]{*}{\shortstack[c]{\(T_\text{process}\)\\ (sec)}}
& \multirow{2}[2]{*}{\shortstack[c]{Frame rate\\ (frames/sec)}}
& UCF-Crime & \multicolumn{2}{c}{XD-Violence}
\\
\cmidrule(lr){7-7}
\cmidrule(lr){8-9}
&&&
&
&
& AUC (\%) & AP (\%) & AUC (\%)
\\
\midrule
1.0 & 0.5 & 8
& {\color{Green} 0.5}
& {\color{Green} 0.364}
& \textbf{82.42}
& 84.66 & 74.15 & 90.43
\\
1.0 & 0.5 & 16
& {\color{Green} 0.5}
& {\color{red} 0.683}
& 44.06
& \textbf{87.33} & \textbf{75.13} & \textbf{90.69}
\\
\rowcolor{mygray}
1.0 & 0.0 & 16
& {\color{Green} 1.0}
& {\color{Green} 0.713}
& 42.06
& 87.29 & 75.13 & 90.54
\\
0.5 & 0.0 & 8
& {\color{Green} 0.5}
& {\color{Green} 0.443}
& 33.88
& 85.56
& 73.81 & 90.52
\\
0.5 & 0.0 & 16
& {\color{Green} 0.5}
& {\color{red} 0.666}
& 22.52
& 85.51 & 69.56 & 88.26
\\
\bottomrule
\end{tabular}
}
}
\hfill

\vspace{-1.6em}

\label{tab:abl}
\end{table*}

\tabref{abl} disentangles the effect of each design choice on accuracy—AUC for UCF-Crime~\cite{Sultani_2018_CVPR} (UCF), AUC and AP for XD-Violence~\cite{Wu2020not} (XD)—and on runtime speed, reported as FPS for 30-fps input.

\textbf{(a) Stability and reproducibility.} We randomly draw four disjoint subsets of 100k captions from the 1M-entry memory and retrain nothing.
The frame-level AUC varies only \(84.38\pm0.60\) on UCF and \(90.13\pm0.07\) on XD, showing that performance does not depend on a particular subset.

\begin{wrapfigure}{R}{0.38\linewidth}
    \centering
    \vspace{-.7em}
    \includegraphics[width=\linewidth]{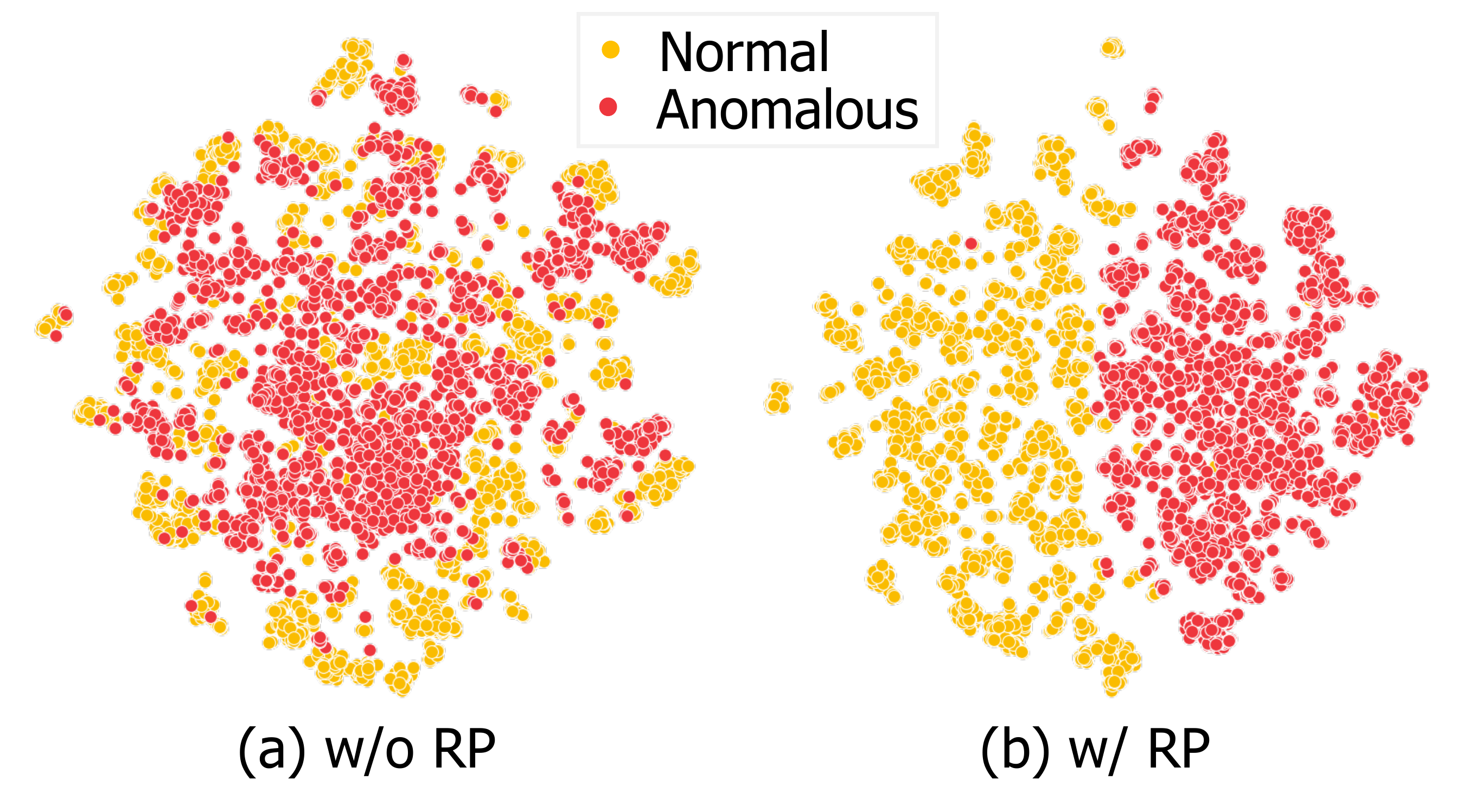}
    \vspace{-0.7em}
    \figcaption{T-SNE embeddings of caption features}{
    We subsample 5,000 normal-anomalous caption pairs
    and visualize (a) before and (b) after applying repulsive prompting (RP).
    RP clearly separates the two groups.
    }
    \figcapmargin
    \label{fig:prior}
\end{wrapfigure}

\textbf{(b) Repulsive prompting (RP).}
Removing RP lowers UCF AUC from 87.29 to 74.98 and XD AP from 75.13 to 71.01.
Meanwhile, the centroid angle shrinks from 33.29° to 8.12°, showing that embeddings collapse without RP as illustrated in \figref{prior}.
We dissect RP into two partial variants:
\textit{keyword-only}, which inserts the tokens \texttt{Normal}/\texttt{Anomalous} but omits the wrapper, and
\textit{template-only}, which adds the wrapper without those tokens.
Both halves recover part of the lost accuracy, yet only their combination restores the full gain.

We also test a purely geometric alternative that projects each segment embedding away from the anomaly axis (details in the supplement).
The tweak gives a small gain on UCF but hurts XD, whereas RP improves both, suggesting that input-level cues are more reliable than post-hoc vector shifts.

\textbf{(c) Top-\(K\) captions.}
We sweep \(K\in\{1,5,10,20,40\}\).
UCF AUC rises from 82.00 at \(K=1\) to 87.29 at \(K=10\) and then drops to 86.31 at \(K=40\); XD shows the same trend. We therefore fix \(K=10\) for all results.
With \(K=40\), many retrieved captions are loosely related to the segment, diluting the soft label mix.
We therefore fix \(K=10\) for all main results.

\textbf{(d) Memory size and throughput.}
Scaling the size of memory from 10k to 1M captions raises UCF AUC from 82.61 to 87.29 and XD AP from 72.17 to 75.13, while fps changes only from 42.95 to 42.06.
Though the performance growth does not saturate, we stop at 1M due to time and cost limits.
As discussed in \secref{comp}, we note that one can construct a larger memory for performance while leaving throughput virtually unchanged as the memory size does not affect the throughput too much.

\begin{wrapfigure}{R}{0.38\linewidth}
    \centering
    \vspace{-0.6em}
    \includegraphics[width=\linewidth]{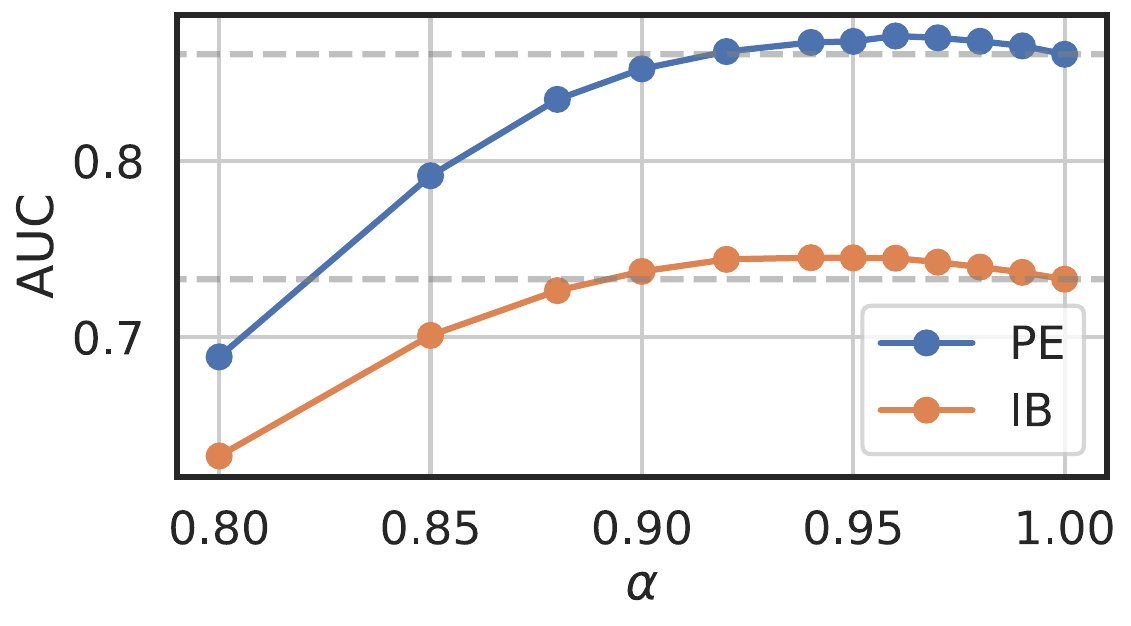}
    \vspace{-1.2em}
    \figcaption{AUC vs. scale factor \(\bm\alpha\)}{
    A mild reduction (\(\alpha\!\approx\!0.95\)) yields favorable AUC, confirming that scaled anomaly penalization is effective without fine-tuning.
    }
    \figcapmargin
    \label{fig:sap}
\end{wrapfigure}

\textbf{(e) Segment length and sampling rate.}
Using 16 frames in a 1s segment reaches the highest scores—87.29 UCF AUC and 75.13 XD AP—but overlapping segments (\(T_{\text{overlap}}=0.5\)s) halves throughput and pushes \(T_{\text{process}}\) beyond the \(T_\text{decision}=0.5\)s limit.
Removing the overlap keeps almost the same accuracy while restoring 42 fps and meeting latency, so the gray row becomes our default.
Shorter windows or eight-frame samples lower UCF AUC to 85.56 and XD AUC to 88.26 yet do not speed the pipeline, offering no benefit.

\textbf{Effect of scaled anomaly penalization (SAP) and choice of \(\bm\alpha\).}
To assess the impact of scaled anomaly penalization (SAP) under realistic conditions, we merge UCF-Crime and XD-Violence into a single evaluation pool and sweep the scale factor \(\alpha\) from 0.80 to 1.00 for two encoders: PerceptionEncoder~\cite{bolya2025PerceptionEncoder} (PE) and ImageBind~\cite{girdhar2023imagebind} (IB). Both achieve peak AUC at \(\alpha\approx0.95\), with stable performance across 0.90-1.0. We therefore fix \(\alpha=0.95\) in all experiments, reducing anomaly bias without per-dataset tuning.

\vspace{-0.3em}
\subsection{Runtime cost}
\label{sec:runtime}

All caption embeddings are computed offline once.
The captions and their embeddings take only 0.3 GiB and 3.9 GiB of memory.
Online inference requires one forward pass of the frozen video encoder per segment and a million dot products.
On a single consumer-grade GPU---42.06 fps on an RTX 3090 and 63.16 fps on an L40S---\ours{} comfortably exceeds the \(\sim\)30 fps real-time threshold.

\section{Conclusions}
\label{sec:conclusions}

We demonstrate that casting video anomaly detection as a caption‐retrieval task can simultaneously achieve zero‐shot deployment, real‐time processing, and interpretable textual explanations.
By building the pseudo‐scene memory offline, we remove all heavy language model inference from the online loop, and by applying repulsive prompting and scaled anomaly penalization, we enforce a clear margin between normal and anomalous captions.
Our method outperforms prior zero‐shot approaches and even several weakly‐supervised baselines, while qualitative analysis shows that retrieved captions align closely with visual evidence.
Ablation studies confirm that (i) the caption construction pipeline is robust, (ii) repulsive prompting consistently improves separation, and (iii) performance is insensitive to the choice of \(\alpha\).
Future work will explore long-range temporal reasoning within this retrieval framework and the incorporation of audio descriptions to enrich the memory.

\section{Limiations}
\label{sec:limit}

Although \ours{} delivers strong zero-shot, real-time results, several limitations remain.
First, our separation of normal and anomalous captions relies on a handcrafted prompt.
A light fine-tuning step on the text encoder—or other debiasing strategies—might widen the margin further, but we have not explored that direction.
Second, the pseudo-scene memory assigns each action a fixed label, yet many behaviors are anomalous only in specific contexts~\cite{cho2024towards}; our current design cannot adapt those labels on-the-fly.
Third, the explainability is constrained to returning the top-matching caption. The system cannot answer open-ended follow-up questions about how severe the anomaly is.
One potential negative impact of our work is that reliance on pseudo‐scene memory may encode and perpetuate biases present in the language model, leading to unfair or inaccurate anomaly detection in certain contexts.
Future work could integrate an unbiased reasoning layer such as a small LLM.

{
    \small
    \bibliographystyle{ieeenat_fullname}
    \bibliography{main}

\begin{thebibliography}{63}
\providecommand{\natexlab}[1]{#1}
\providecommand{\url}[1]{\texttt{#1}}
\expandafter\ifx\csname urlstyle\endcsname\relax
  \providecommand{\doi}[1]{doi: #1}\else
  \providecommand{\doi}{doi: \begingroup \urlstyle{rm}\Url}\fi

\bibitem[Ahn et~al.(2025)Ahn, Jo, Lee, Kwon, Hong, and Park]{ahn2025anyanomaly}
S. Ahn, Y. Jo, K. Lee, S. Kwon, I. Hong, and S. Park.
\newblock Anyanomaly: Zero-shot customizable video anomaly detection with lvlm.
\newblock arXiv preprint arXiv:2503.04504, 2025.

\bibitem[Alayrac et~al.(2022)Alayrac, Donahue, Luc, Miech, Barr, Hasson, Lenc, Mensch, Zisserman, and Simonyan]{alayrac2022flamingo}
Jean-Baptiste Alayrac, Jeff Donahue, Pauline Luc, Antoine Miech, Iain Barr, Yana Hasson, Karel Lenc, Arthur Mensch, Andrew Zisserman, and Karen Simonyan.
\newblock Flamingo: a visual language model for few-shot learning.
\newblock In \emph{NeurIPS}, 2022.

\bibitem[Antol et~al.(2015)Antol, Agrawal, Lu, Mitchell, Batra, Parikh, and Zitnick]{VQA}
Stanislaw Antol, Aishwarya Agrawal, Jiasen Lu, Margaret Mitchell, Dhruv Batra, Devi Parikh, and C.~Lawrence Zitnick.
\newblock Vqa: Visual question answering.
\newblock In \emph{ICCV}, 2015.

\bibitem[Bar(2007)]{Bar2007}
Moshe Bar.
\newblock The proactive brain: Using analogies and associations to generate predictions.
\newblock \emph{Trends in Cognitive Sciences}, 2007.

\bibitem[Bogdoll et~al.(2022)Bogdoll, Nitsche, and Z\"ollner]{Bogdoll_2022_CVPR}
Daniel Bogdoll, Maximilian Nitsche, and J.~Marius Z\"ollner.
\newblock Anomaly detection in autonomous driving: A survey.
\newblock In \emph{IEEE Conf. Comput. Vis. Pattern Recog. Worksh.}, 2022.

\bibitem[Bolya et~al.(2025)Bolya, Huang, Sun, Cho, Madotto, Wei, Ma, Zhi, Rajasegaran, Rasheed, Wang, Monteiro, Xu, Dong, Ravi, Li, Doll{\'a}r, and Feichtenhofer]{bolya2025PerceptionEncoder}
Daniel Bolya, Po-Yao Huang, Peize Sun, Jang~Hyun Cho, Andrea Madotto, Chen Wei, Tengyu Ma, Jiale Zhi, Jathushan Rajasegaran, Hanoona Rasheed, Junke Wang, Marco Monteiro, Hu Xu, Shiyu Dong, Nikhila Ravi, Daniel Li, Piotr Doll{\'a}r, and Christoph Feichtenhofer.
\newblock Perception encoder: The best visual embeddings are not at the output of the network.
\newblock \emph{arXiv:2504.13181}, 2025.

\bibitem[Brown et~al.(2020)Brown, Mann, Ryder, Subbiah, Kaplan, Dhariwal, Neelakantan, et~al.]{brown2020gpt3}
Tom~B. Brown, Benjamin Mann, Nick Ryder, Melanie Subbiah, Jared Kaplan, Prafulla Dhariwal, Arvind Neelakantan, et~al.
\newblock Language models are few-shot learners.
\newblock In \emph{NeurIPS}, 2020.

\bibitem[{BusinessWire}(2023)]{businesswire2023surveillance}
{BusinessWire}.
\newblock {Global Surveillance Camera Market Poised to Reach US\$33.11 Billion by 2023; 278.6 Million Units Forecast}.
\newblock \url{https://www.businesswire.com/news/home/20231025926921/en/Global-Surveillance-Camera-Market-Poised-to-Reach-US33.11-Billion-by-2023}, 2023.
\newblock Accessed: May 2025.

\bibitem[Chen et~al.(2023)Chen, Liu, Zhang, Fok, Qi, and Wu]{10.1609/aaai.v37i1.25112}
Yingxian Chen, Zhengzhe Liu, Baoheng Zhang, Wilton Fok, Xiaojuan Qi, and Yik-Chung Wu.
\newblock Mgfn: magnitude-contrastive glance-and-focus network for weakly-supervised video anomaly detection.
\newblock In \emph{AAAI}, 2023.

\bibitem[Cho et~al.(2024)Cho, Kim, Shim, Wee, and Lee]{cho2024towards}
MyeongAh Cho, Taeoh Kim, Minho Shim, Dongyoon Wee, and Sangyoun Lee.
\newblock Towards multi-domain learning for generalizable video anomaly detection.
\newblock In \emph{NeurIPS}, 2024.

\bibitem[Clark and Etzioni(2018)]{clark2018arc}
Peter Clark and Oren Etzioni.
\newblock Think you have solved the ai2 reasoning challenge? reconsidering the arc dataset.
\newblock In \emph{ACL}, 2018.

\bibitem[Dai et~al.(2023)Dai, Li, Li, Tiong, et~al.]{dai2023instructblip}
Wenliang Dai, Junnan Li, Dongxu Li, Anthony M.~H. Tiong, et~al.
\newblock Instructblip: Towards general-purpose vision-language models with instruction tuning.
\newblock In \emph{NeurIPS}, 2023.

\bibitem[Dev et~al.(2024)Dev, Hazari, and Das]{prasad2024mca}
Prabhu~Prasad Dev, Raju Hazari, and Pranesh Das.
\newblock Mcanet: Multimodal caption aware training-free video anomaly detection via large language model.
\newblock In \emph{ICPR}, 2024.

\bibitem[Friston(2005)]{Friston2005}
Karl Friston.
\newblock A theory of cortical responses.
\newblock \emph{Philosophical Transactions of the Royal Society~B}, 2005.

\bibitem[Girdhar et~al.(2023)Girdhar, El-Nouby, Liu, Singh, Alwala, Joulin, and Misra]{girdhar2023imagebind}
Rohit Girdhar, Alaaeldin El-Nouby, Zhuang Liu, Mannat Singh, Kalyan~Vasudev Alwala, Armand Joulin, and Ishan Misra.
\newblock Imagebind: One embedding space to bind them all.
\newblock In \emph{CVPR}, 2023.

\bibitem[Gong and \emph{et al.}(2024)]{gong2024anomalyclip}
H. Gong and \emph{et al.}
\newblock Anomalyclip: Object-agnostic prompt learning for zero-shot anomaly detection.
\newblock In \emph{ICLR}, 2024.

\bibitem[Goyal et~al.(2017)Goyal, Khot, Summers{-}Stay, Batra, and Parikh]{VQAv2}
Yash Goyal, Tejas Khot, Douglas Summers{-}Stay, Dhruv Batra, and Devi Parikh.
\newblock Making the {V} in {VQA} matter: Elevating the role of image understanding in visual question answering.
\newblock In \emph{CVPR}, 2017.

\bibitem[{Grand View Research}(2024)]{grandview2024video}
{Grand View Research}.
\newblock {Video Surveillance Market Size, Share \& Trends Analysis Report By Component, By Deployment Mode, By Application, By Region -- Global Forecast to 2030}.
\newblock \url{https://www.grandviewresearch.com/industry-analysis/video-surveillance-market-report}, 2024.
\newblock Accessed: May 2025.

\bibitem[Hasan et~al.(2016)Hasan, Choi, Neumann, Roy-Chowdhury, and Davis]{Hasan_2016_CVPR}
Mahmudul Hasan, Jonghyun Choi, Jan Neumann, Amit~K. Roy-Chowdhury, and Larry~S. Davis.
\newblock Learning temporal regularity in video sequences.
\newblock In \emph{CVPR}, 2016.

\bibitem[Hudson and Manning(2019)]{GQA}
Drew~A. Hudson and Christopher~D. Manning.
\newblock Gqa: A new dataset for real-world visual reasoning and compositional question answering.
\newblock In \emph{CVPR}, 2019.

\bibitem[Joo et~al.(2023)Joo, Vo, Yamazaki, and Le]{joo2023cliptsa}
Hyekang~Kevin Joo, Khoa Vo, Kashu Yamazaki, and Ngan Le.
\newblock Clip-tsa: Clip-assisted temporal self-attention for weakly-supervised video anomaly detection.
\newblock In \emph{ICIP}, 2023.

\bibitem[Joshi et~al.(2017)Joshi, Choi, Weld, and Zettlemoyer]{joshi2017triviaqa}
Mandar Joshi, Eunsol Choi, Daniel Weld, and Luke Zettlemoyer.
\newblock Triviaqa: A large scale distantly supervised challenge dataset for reading comprehension.
\newblock In \emph{ACL}, 2017.

\bibitem[Karim et~al.(2024)Karim, Pande, and Ahuja]{karim2024reward}
H. Karim, V. Pande, and N. Ahuja.
\newblock Reward: Real-time weakly supervised video anomaly detection.
\newblock In \emph{WACV}, 2024.

\bibitem[Kwiatkowski and Palmer(2019)]{kwiatkowski2019nq}
Tom Kwiatkowski and Alexis et~al. Palmer.
\newblock Natural questions: a benchmark for question answering research.
\newblock \emph{TACL}, 2019.

\bibitem[Li et~al.(2023)Li, Li, Savarese, and Hoi]{li2023blip}
Junnan Li, Dongxu Li, Silvio Savarese, and Steven Hoi.
\newblock Blip-2: Bootstrapping language-image pre-training with frozen image encoders and large language models.
\newblock In \emph{International conference on machine learning}, pages 19730--19742. PMLR, 2023.

\bibitem[Li et~al.(2022)Li, Liu, and Jiao]{Li_Liu_Jiao_2022}
Shuo Li, Fang Liu, and Licheng Jiao.
\newblock Self-training multi-sequence learning with transformer for weakly supervised video anomaly detection.
\newblock In \emph{AAAI}, 2022.

\bibitem[Li et~al.(2024)Li, Lin, and Pei]{li2024multi}
Shengzhi Li, Rongyu Lin, and Shichao Pei.
\newblock Multi-modal preference alignment remedies degradation of visual instruction tuning on language models.
\newblock \emph{arXiv preprint arXiv:2402.10884}, 2024.

\bibitem[Liu et~al.(2023{\natexlab{a}})Liu, Li, Li, and Lee]{liu2023improvedllava}
Haotian Liu, Chunyuan Li, Yuheng Li, and Yong~Jae Lee.
\newblock Improved baselines with visual instruction tuning, 2023{\natexlab{a}}.

\bibitem[Liu et~al.(2023{\natexlab{b}})Liu, Li, Wu, and Lee]{liu2023llava}
Haotian Liu, Chunyuan Li, Qingyang Wu, and Yong~Jae Lee.
\newblock Visual instruction tuning.
\newblock In \emph{NeurIPS}, 2023{\natexlab{b}}.
\newblock Oral.

\bibitem[Lu et~al.(2013)Lu, Shi, and Jia]{Lu_2013_ICCV}
Cewu Lu, Jianping Shi, and Jiaya Jia.
\newblock Abnormal event detection at 150 fps in matlab.
\newblock In \emph{ICCV}, 2013.

\bibitem[Marino et~al.(2019)Marino, Yu, Zhang, Luo, Bansal, Lee, and Batra]{OKVQA}
Kenneth Marino, Zhou Yu, Yuchen Zhang, Junjie Luo, Mohit Bansal, Stefan Lee, and Dhruv Batra.
\newblock {OK{-}VQA}: A visual question answering benchmark requiring external knowledge.
\newblock In \emph{CVPR}, 2019.

\bibitem[Micorek et~al.(2024)Micorek, Rudzinski, and Zhang]{micorek2024mulde}
M. Micorek, M. Rudzinski, and L. Zhang.
\newblock Mulde: Multi-scale log-density estimation for video anomaly detection.
\newblock In \emph{CVPR}, 2024.

\bibitem[OpenAI(2024)]{openai_gpt4o}
OpenAI.
\newblock Gpt-4o: Openai's omnimodal model.
\newblock \url{https://openai.com/index/gpt-4o}, 2024.

\bibitem[Ouyang et~al.(2022)Ouyang, Wu, Jiang, Almeida, Wainwright, Mishkin, Zhang, Agarwal, et~al.]{ouyang2022instructgpt}
Long Ouyang, Jeff Wu, Xu Jiang, Diogo Almeida, Carroll~L. Wainwright, Pamela Mishkin, Chong Zhang, Sandhini Agarwal, et~al.
\newblock Training language models to follow instructions with human feedback.
\newblock In \emph{NeurIPS}, 2022.

\bibitem[Paperno et~al.(2016)Paperno, Kruszewski, Lazaridou, and Baroni]{paperno2016lambada}
Denis Paperno, Germ\'an Kruszewski, Angeliki Lazaridou, and Marco Baroni.
\newblock The lambada dataset: Word prediction requiring a broad discourse context.
\newblock In \emph{ACL}, 2016.

\bibitem[Peng et~al.(2024)Peng, Wang, Dong, Hao, Huang, Ma, Ye, and Wei]{peng2024kosmos2}
Zhiliang Peng, Wenhui Wang, Li Dong, Yaru Hao, Shaohan Huang, Shuming Ma, Qixiang Ye, and Furu Wei.
\newblock Grounding multimodal large language models to the world.
\newblock In \emph{ICLR}, 2024.

\bibitem[Roth et~al.(2022)Roth, Pemula, Zepeda, Sch\"olkopf, Brox, and Gehler]{Roth_2022_CVPR}
Karsten Roth, Latha Pemula, Joaquin Zepeda, Bernhard Sch\"olkopf, Thomas Brox, and Peter Gehler.
\newblock Towards total recall in industrial anomaly detection.
\newblock In \emph{CVPR}, 2022.

\bibitem[Shao et~al.(2025)Shao, He, Li, Chen, Long, Zeng, Fan, Zhang, Yan, Ma, et~al.]{shao2025eventvad}
Yihua Shao, Haojin He, Sijie Li, Siyu Chen, Xinwei Long, Fanhu Zeng, Yuxuan Fan, Muyang Zhang, Ziyang Yan, Ao Ma, et~al.
\newblock Eventvad: Training-free event-aware video anomaly detection.
\newblock \emph{arXiv preprint arXiv:2504.13092}, 2025.

\bibitem[Sultani et~al.(2018)Sultani, Chen, and Shah]{Sultani_2018_CVPR}
Waqas Sultani, Chen Chen, and Mubarak Shah.
\newblock Real-world anomaly detection in surveillance videos.
\newblock In \emph{CVPR}, 2018.

\bibitem[Summerfield and de~Lange(2014)]{Summerfield2014}
Christopher Summerfield and Floris~P. de Lange.
\newblock Expectation in perceptual decision making: Neural and computational mechanisms.
\newblock \emph{Nature Reviews Neuroscience}, 2014.

\bibitem[Talmor et~al.(2019)Talmor, Herzig, Lourie, and Berant]{talmor2019commonsenseqa}
Alon Talmor, Jonathan Herzig, Nicholas Lourie, and Jonathan Berant.
\newblock Commonsenseqa: A question answering challenge targeting commonsense knowledge.
\newblock In \emph{NACCL}, 2019.

\bibitem[Thakare et~al.(2023{\natexlab{a}})Thakare, Dogra, Choi, Kim, and Kim]{Thakare_2023_rareanom}
Kamalakar~Vijay Thakare, Debi~Prosad Dogra, Heeseung Choi, Haksub Kim, and Ig-Jae Kim.
\newblock Rareanom: A benchmark video dataset for rare type anomalies.
\newblock \emph{PR}, 2023{\natexlab{a}}.

\bibitem[Thakare et~al.(2023{\natexlab{b}})Thakare, Raghuwanshi, Dogra, Choi, and Kim]{Thakare_2023_WACV}
Kamalakar~Vijay Thakare, Yash Raghuwanshi, Debi~Prosad Dogra, Heeseung Choi, and Ig-Jae Kim.
\newblock Dyannet: A scene dynamicity guided self-trained video anomaly detection network.
\newblock In \emph{WACV}, 2023{\natexlab{b}}.

\bibitem[Tian et~al.(2021)Tian, Pang, Chen, Singh, Verjans, and Carneiro]{Tian_2021_ICCV}
Yu Tian, Guansong Pang, Yuanhong Chen, Rajvinder Singh, Johan~W. Verjans, and Gustavo Carneiro.
\newblock Weakly-supervised video anomaly detection with robust temporal feature magnitude learning.
\newblock In \emph{ICCV}, 2021.

\bibitem[Touvron et~al.(2023)Touvron, Martin, Stone, Barulina, Borlaug, Azhar, Dror, Joulin, Grave, and Conneau]{touvron2023llama2}
Hugo Touvron, Louis Martin, Kevin Stone, Petr Barulina, Kevin Borlaug, Faisal Azhar, Gideon Dror, Armand Joulin, Edouard Grave, and Alexis Conneau.
\newblock {LLaMA 2}: Open foundation and fine‐tuned language models.
\newblock \emph{arXiv preprint arXiv:2307.09288}, 2023.

\bibitem[Tur et~al.(2023{\natexlab{a}})Tur, Dall'Asen, Beyan, and Ricci]{tur2023diffusion}
Anil~Osman Tur, Nicola Dall'Asen, Cigdem Beyan, and Elisa Ricci.
\newblock Unsupervised video anomaly detection with diffusion models conditioned on compact motion representations.
\newblock In \emph{ICIAP}, 2023{\natexlab{a}}.

\bibitem[Tur et~al.(2023{\natexlab{b}})Tur, Dall’Asen, Beyan, and Ricci]{tur2023exploring}
Anil~Osman Tur, Nicola Dall’Asen, Cigdem Beyan, and Elisa Ricci.
\newblock Exploring diffusion models for unsupervised video anomaly detection.
\newblock In \emph{ICIP}, 2023{\natexlab{b}}.

\bibitem[Wang and Cherian(2019)]{Wang_2019_ICCV}
Jue Wang and Anoop Cherian.
\newblock Gods: Generalized one-class discriminative subspaces for anomaly detection.
\newblock In \emph{ICCV}, 2019.

\bibitem[Wang et~al.(2023)Wang, Wei, Wang, Schuurmans, Bosma, Chi, Le, Zhou, et~al.]{wang2023selfinstruct}
Yizhong Wang, Jason Wei, Xuezhi Wang, Dale Schuurmans, Maarten Bosma, Ed~H. Chi, Quoc~V. Le, Denny Zhou, et~al.
\newblock Self-instruct: Aligning language models with self generated instructions.
\newblock In \emph{ACL}, 2023.

\bibitem[Wei et~al.(2022{\natexlab{a}})Wei, Wang, Schuurmans, Bosma, Ichter, Le, and Chi]{wei2022cot}
Jason Wei, Xuezhi Wang, Dale Schuurmans, Maarten Bosma, Brian Ichter, Quoc~V. Le, and Ed~H. Chi.
\newblock Chain-of-thought prompting elicits reasoning in large language models.
\newblock In \emph{NeurIPS}, 2022{\natexlab{a}}.

\bibitem[Wei et~al.(2022{\natexlab{b}})Wei, Wang, Schuurmans, Bosma, Ichter, Le, and Chi]{wei2022flan}
Jason Wei, Xuezhi Wang, Dale Schuurmans, Maarten Bosma, Brian Ichter, Quoc~V. Le, and Ed~H. Chi.
\newblock Finetuned language models are zero-shot learners.
\newblock In \emph{ICLR}, 2022{\natexlab{b}}.

\bibitem[Wu et~al.(2022)Wu, Hsieh, Chen, Fuh, and Liu]{WuHCFL22}
Jhih-Ciang Wu, He-Yen Hsieh, Ding-Jie Chen, Chiou-Shann Fuh, and Tyng-Luh Liu.
\newblock Self-supervised sparse representation for video anomaly detection.
\newblock In \emph{ECCV}, 2022.

\bibitem[Wu and Liu(2021)]{10.1109/TIP.2021.3062192}
Peng Wu and Jing Liu.
\newblock Learning causal temporal relation and feature discrimination for anomaly detection.
\newblock \emph{TIP}, 2021.

\bibitem[Wu et~al.(2020)Wu, Liu, Shi, Sun, Shao, Wu, and Yang]{Wu2020not}
Peng Wu, jing Liu, Yujia Shi, Yujia Sun, Fangtao Shao, Zhaoyang Wu, and Zhiwei Yang.
\newblock Not only look, but also listen: Learning multimodal violence detection under weak supervision.
\newblock In \emph{ECCV}, 2020.

\bibitem[Wu et~al.(2024)Wu, Zhou, Pang, Zhou, Yan, Wang, and Zhang]{wu2023vadclip}
Peng Wu, Xuerong Zhou, Guansong Pang, Lingru Zhou, Qingsen Yan, Peng Wang, and Yanning Zhang.
\newblock Vadclip: Adapting vision-language models for weakly supervised video anomaly detection.
\newblock \emph{AAAI}, 2024.

\bibitem[Wu et~al.(2025)Wu, Su, Pang, Sun, Yan, Wang, and Zhang]{wu2025avadclip}
Peng Wu, Wanshun Su, Guansong Pang, Yujia Sun, Qingsen Yan, Peng Wang, and Yanning Zhang.
\newblock Avadclip: Audio-visual collaboration for robust video anomaly detection.
\newblock \emph{arXiv preprint arXiv:2504.04495}, 2025.

\bibitem[Wu and \emph{et al.}(2024)]{wu2024openvocabulary}
Y. Wu and \emph{et al.}
\newblock Open-vocabulary video anomaly detection.
\newblock In \emph{CVPR}, 2024.

\bibitem[Ye et~al.(2024)Ye, Liu, and He]{ye2024vera}
Muchao Ye, Weiyang Liu, and Pan He.
\newblock Vera: Explainable video anomaly detection via verbalized learning of vision-language models.
\newblock \emph{arXiv preprint arXiv:2412.01095}, 2024.

\bibitem[Zaheer et~al.(2022)Zaheer, Mahmood, Khan, Segu, Yu, and Lee]{Zaheer_2022_CVPR}
M.~Zaigham Zaheer, Arif Mahmood, M.~Haris Khan, Mattia Segu, Fisher Yu, and Seung-Ik Lee.
\newblock Generative cooperative learning for unsupervised video anomaly detection.
\newblock In \emph{CVPR}, 2022.

\bibitem[Zanella et~al.(2024)Zanella, Menapace, Mancini, Wang, and Ricci]{zanella2024harnessing}
Luca Zanella, Willi Menapace, Massimiliano Mancini, Yiming Wang, and Elisa Ricci.
\newblock Harnessing large language models for training-free video anomaly detection.
\newblock In \emph{CVPR}, 2024.

\bibitem[Zhang et~al.(2024)Zhang, Xu, Wang, Zuo, Han, Huang, Gao, Wang, and Sang]{zhang2024holmes}
Huaxin Zhang, Xiaohao Xu, Xiang Wang, Jialong Zuo, Chuchu Han, Xiaonan Huang, Changxin Gao, Yuehuan Wang, and Nong Sang.
\newblock Holmes-vad: Towards unbiased and explainable video anomaly detection via multi-modal llm.
\newblock \emph{arXiv preprint arXiv:2406.12235}, 2024.

\bibitem[Zhang et~al.(2019)Zhang, Qing, and Miao]{zhang2019temporal}
Jiangong Zhang, Laiyun Qing, and Jun Miao.
\newblock Temporal convolutional network with complementary inner bag loss for weakly supervised anomaly detection.
\newblock In \emph{ICIP}, 2019.

\bibitem[Zhu et~al.(2021)Zhu, Chen, and Sultani]{zhu2021video}
Sijie Zhu, Chen Chen, and Waqas Sultani.
\newblock Video anomaly detection for smart surveillance.
\newblock In \emph{Computer Vision: A Reference Guide}. Springer, 2021.

\end{thebibliography}
}

\end{document}